\documentclass[letterpaper]{article}

\PassOptionsToPackage{numbers}{natbib}
\usepackage[preprint]{nips_2018} % arxiv
\usepackage[utf8]{inputenc} % allow utf-8 input
\usepackage[T1]{fontenc}    % use 8-bit T1 fonts
\usepackage{hyperref}       % hyperlinks
\usepackage{url}            % simple URL typesetting
\usepackage{booktabs}       % professional-quality tables
\usepackage{amsfonts}       % blackboard math symbols
\usepackage{nicefrac}       % compact symbols for 1/2
\usepackage{microtype}      % microtypography
\usepackage{amsmath}
\usepackage{algorithm, algorithmicx, algpseudocode}
\usepackage[pdftex]{graphicx} % for arxiv
\usepackage{comment}
\usepackage{color}
\usepackage{bmpsize}
\usepackage{subcaption}
\usepackage{enumitem}

\title{Tile2Vec: Unsupervised representation learning for spatially distributed data}

\author{
  Neal Jean \\
  Stanford University \\
  \texttt{nealjean@stanford.edu} \\
  \And
  Sherrie Wang \\
  Stanford University \\
  \texttt{sherwang@stanford.edu} \\
  \And
  Anshul Samar \\
  Stanford University \\
  \texttt{asamar@stanford.edu} \\
  \AND
  George Azzari \\
  Stanford University \\
  \texttt{gazzari@stanford.edu} \\
  \And
  David Lobell \\
  Stanford University \\
  \texttt{dlobell@stanford.edu} \\
  \And
  Stefano Ermon \\
  Stanford University \\
  \texttt{ermon@stanford.edu} \\
}

\begin{document}

\maketitle

%%%%%%%%%%%%%%%%%%%%%%%%
% ABSTRACT
%%%%%%%%%%%%%%%%%%%%%%%%

\begin{abstract}
Geospatial analysis lacks methods like the word vector representations and pre-trained networks that significantly boost performance across a wide range of natural language and computer vision tasks. To fill this gap, we introduce Tile2Vec, an unsupervised representation learning algorithm that extends the distributional hypothesis from natural language --- words appearing in similar contexts tend to have similar meanings --- to spatially distributed data. We demonstrate empirically that Tile2Vec learns semantically meaningful representations on three datasets. Our learned representations significantly improve performance in downstream classification tasks and, similar to word vectors, visual analogies can be obtained via simple arithmetic in the latent space.
\end{abstract}

%%%%%%%%%%%%%%%%%%%%%%%%
%%%%%%%%%%%%%%%%%%%%%%%%

\section{Introduction}

Remote sensing, the measurement of the Earth's surface through aircraft- or satellite-based sensors, is becoming increasingly important to many applications, including land use monitoring, precision agriculture, and military intelligence \cite{foody,mulla,madden2009manual}.
Combined with recent advances in deep learning and computer vision \cite{krizhevsky2012imagenet,he2016deep}, there is enormous potential for monitoring global issues through the automated analysis of remote sensing and other geospatial data streams.
However, recent successes in machine learning have largely relied on supervised learning techniques and the availability of very large annotated datasets.
Remote sensing provides a huge supply of data, but many downstream tasks of interest are constrained by a lack of labels.

The research community has developed a number of techniques to mitigate the need for labeled data.
Often, the key underlying idea is to find a low-dimensional \emph{representation} of the data that is more suitable for downstream machine learning tasks. In many NLP applications, pre-trained word vectors have led to dramatic performance improvements. In computer vision, pre-training on ImageNet is a \emph{de facto} standard that drastically reduces the amount of training data needed for new tasks. Existing techniques, however, are not suitable for remote sensing data that, while superficially resembling natural images, have unique characteristics that require new methodologies.
Unlike natural images --- object-centric, two-dimensional depictions of three-dimensional scenes --- remote sensing images are taken from a bird's eye perspective and are often \emph{multi-spectral}, presenting both challenges and opportunities. On one hand, models pre-trained on ImageNet do not perform well and cannot take advantage of additional spectral bands. On the other, there are fewer occlusions, permutations of object placement, and changes of scale to contend with --- this spatial coherence provides a powerful signal for learning representations.

Our main assumption is that image tiles that are geographic neighbors (i.e. close spatially) should have similar semantics and therefore representations, while tiles far apart are likely to have dissimilar semantics and should therefore have dissimilar representations. This is akin to the \emph{distributional hypothesis} used to construct word vector representations in natural language: words that appear in similar contexts should have similar meanings. The main computational (and statistical) challenge is that image patches are themselves complex, high-dimensional vectors, unlike words.

In this paper, we propose Tile2Vec, a method for learning compressed yet informative representations from unlabeled remote sensing data.
We evaluate our algorithm on a wide range of remote sensing datasets and find that it generalizes across data modalities, with stable training and robustness to hyperparameter choices.
On a difficult land use classification task, our learned representations outperform other unsupervised features and even exceed the performance of supervised models trained on large labeled training sets.
Tile2Vec representations also reside in a meaningful embedding space, demonstrated through visual query by example and visual analogy experiments.
Finally, we apply Tile2Vec to the non-image task of predicting country health indices from economic data, suggesting that real-world applications of Tile2Vec may extend to domains beyond remote sensing.

\section{Tile2Vec} \label{tile2vec}

\subsection{Distributional semantics}

The distributional hypothesis in linguistics is the idea that ``a word is characterized by the company it keeps''.
In NLP, algorithms like Word2vec and GloVe leverage this assumption to learn continuous representations that capture the nuanced meanings of huge vocabularies of words.
The strategy is to build a co-occurrence matrix and learn a low-rank approximation in which words that appear in similar contexts have similar representations \cite{levy2014neural,pennington2014glove,mikolov2013distributed}.

To extend these ideas to geospatial data, we need to answer the following questions:
\vspace{-0.2cm}
\begin{itemize}
\item What is the right atomic unit, i.e., the equivalent of individual words in NLP?
\vspace{-0.1cm}
\item What is the right notion of context?
\end{itemize}
\vspace{-0.1cm}
For atomic units, we propose to learn representations at the level of remote sensing \emph{tiles}, a generalization of image patches to multi-spectral data.
This introduces new challenges as tiles are high-dimensional objects --- computations on co-occurrence matrices of tiles would quickly become intractable, and statistics almost impossible to estimate from finite data.
Convolutional neural networks (CNNs) will play a crucial role in projecting down the dimensionality of our inputs.

For context, we rely on spatial \emph{neighborhoods}.
Distance in geographic space provides a form of weak supervision: we assume that tiles that are close together have similar semantics and therefore should, on average, have more similar representations than tiles that are far apart.
By exploiting this fact that landscapes in remote sensing datasets are highly spatially correlated, we hope to extract enough learning signal to reliably train deep neural networks.

%%%%%%% PSEUDOCODE %%%%%%%%
\algrenewcommand\alglinenumber[1]{\tiny #1:}

\begin{figure}
\begin{minipage}{\textwidth}
\centering
\begin{minipage}{.49\linewidth}
\begin{algorithm}[H]
\scriptsize
\caption{SampleTileTriplets($D, N, s, r$)}
\label{alg:sampling}
\begin{algorithmic}[1]
\State \textbf{Input:} Image dataset $D$, number of triplets $N$, tile size $s$, neighborhood radius $r$
\State \textbf{Output:} Tile triplets $T = \{(t_a^{(i)}, t_n^{(i)}, t_d^{(i)})\}_{i=1}^N$ \\
\State Initialize tile triplets $T = \{\}$
% Loop through all triplets
\For {$i \leftarrow 1, N$}
% Sample anchor tile
\State $t_a^{(i)} \leftarrow \textproc{SampleTile}(D, s)$
% Sample neighbor tile
\State $t_n^{(i)} \leftarrow \textproc{SampleTile}(\textproc{Neighborhood}(D, r, t_a^{(i)}), s)$
% Sample distant tile
\State $t_d^{(i)} \leftarrow \textproc{SampleTile}(\neg \textproc{Neighborhood}(D, r, t_a^{(i)}), s)$
% Add triplet to dataset
\State Update $T \leftarrow T \cup (t_a^{(i)}, t_n^{(i)}, t_d^{(i)})$
\EndFor
\State \Return $T$ \\

%SampleTile
\Function{SampleTile}{$A, s$}
\State $t \leftarrow \textrm{Sample tile of size $s$ uniformly at random from $A$}$
\State \Return $t$
\EndFunction \\

%Neighborhood
\Function{Neighborhood}{$D, r, t$}
\State $A \leftarrow \textrm{Subset of $D$ within radius $r$ of tile $t$}$
\State \Return $A$
\EndFunction

\end{algorithmic}
\end{algorithm}
\end{minipage}
\hspace{0.02\linewidth}
\begin{minipage}{.47\textwidth}
\vspace{0.4cm}
\includegraphics[width=\linewidth]{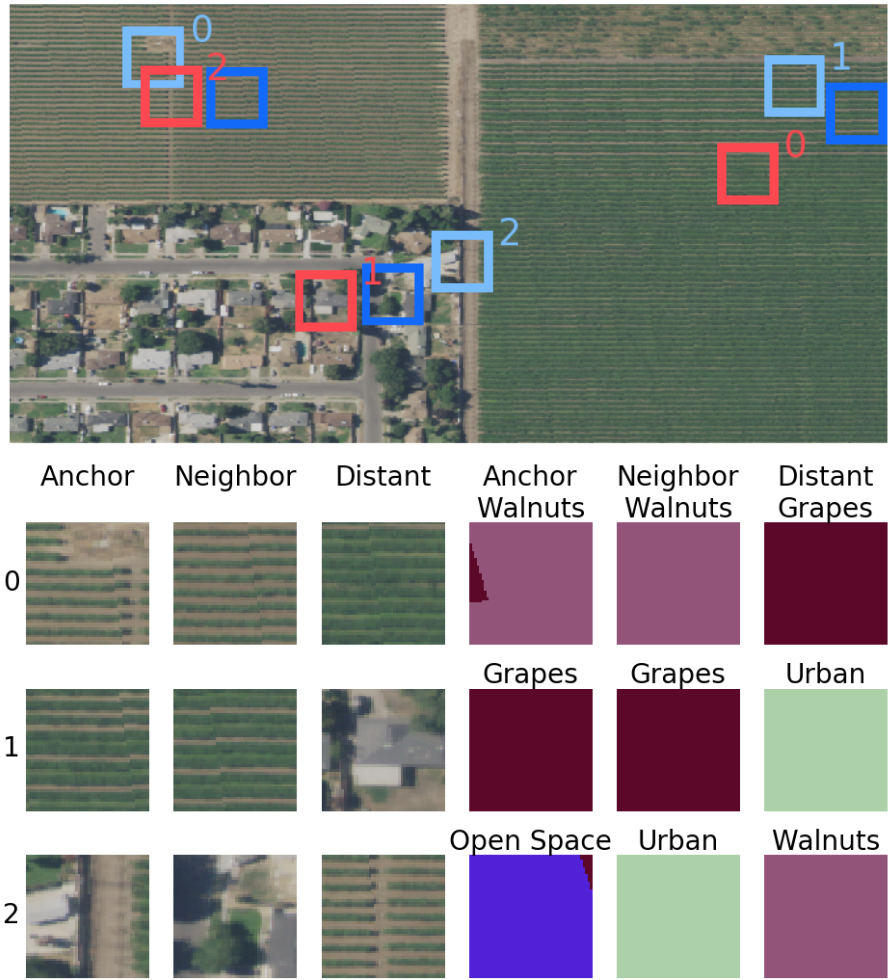}
\end{minipage}

\end{minipage}
\caption{\textbf{Left:} Tile2Vec triplet sampling algorithm. \textbf{Right:} (Top) Light blue boxes denote anchor tiles, dark blue neighbor tiles, and red distant tiles.
(Bottom) Tile triplets corresponding to the top panel. The columns show anchor, neighbor, and distant tiles and their respective CDL class labels. Anchor and neighbor tiles tend to be the same class, while anchor and distant tend to be different.}
\label{fig:sampling}
\end{figure}
%%%%%%%%%%%%%%%%%%%%%%%%%%%

%%%%%%%%%%%%%%%%%%%%%%%%
%%%%%%%%%%%%%%%%%%%%%%%%

\subsection{Unsupervised triplet loss}

To learn a mapping from image tiles to low-dimensional embeddings, we train a convolutional neural network on triplets of tiles, where each triplet consists of an anchor tile $t_{a}$, a neighbor tile $t_{n}$ that is close geographically, and a distant tile $t_{d}$ that is farther away.
Following our distributional assumption, we want to minimize the Euclidean distance between the embeddings of the anchor tile and the neighbor tile, while maximizing the distance between the anchor and distant embeddings.
For each tile triplet $(t_{a}, t_{n}, t_{d})$, we seek to minimize the triplet loss
\begin{align} \label{tripletloss}
L(t_{a}, t_{n}, t_{d}) = \left[ \vert\vert f_{\theta} (t_{a}) - f_{\theta} (t_{n}) \vert\vert_{2} - \vert\vert f_{\theta} (t_{a}) - f_{\theta} (t_{d}) \vert\vert_{2} + m \right]_{+}
\end{align}
To prevent the network from pushing the distant tile farther without restriction, we introduce a rectifier with margin $m$: once the distance to the distant embedding is greater than the distance to the neighbor embedding by at least the margin, we are satisfied.
Here, $f_\theta$ is a CNN with parameters $\theta$ that maps from the domain of image tiles $\mathcal{X}$ to $d$-dimensional real-valued vector representations, $f_\theta: \mathcal{X} \rightarrow \mathbb{R}^d$.

Notice that when $\vert\vert f_{\theta} (t_{a}) - f_{\theta} (t_{n}) \vert\vert_{2} < \vert\vert f_{\theta} (t_{a}) - f_{\theta} (t_{d}) \vert\vert_{2}$, all embeddings can be scaled by some constant in order to satisfy the margin and bring the loss to zero.
We observe this behavior empirically: beyond a small number of iterations, the CNN learns to increase embedding magnitudes and the loss decreases to zero.
By penalizing the embeddings' $l^{2}$-norms, we constrain the network to generate embeddings within a hypersphere, leading to a representation space in which relative distances have meaning.
Given a dataset of $N$ tile triplets, our full training objective is
\begin{align} \label{objective}
\min_\theta \sum_{i=1}^N & \left[ L(t_a^{(i)}, t_n^{(i)}, t_d^{(i)}) + \lambda \left( ||z_a^{(i)}||_2 + ||z_a^{(i)}||_2 + ||z_a^{(i)}||_2 \right) \right],
\end{align}
where $z_a^{(i)} = f_\theta(t_a^{(i)}) \in \mathbb{R}^d$ and similarly for $z_n^{(i)}$ and $z_d^{(i)}$.

%%%%%%%%%%%%%%%%%%%%%%%%

\subsection{Triplet sampling}

The sampling procedure for $t_{a}$, $t_{n}$, and $t_{d}$ is described by two parameters:
\vspace{-0.2cm}
\begin{itemize}
\item \textbf{Tile size} defines the pixel width and height of a single tile.
\item \textbf{Neighborhood} defines the region around the anchor tile from which to sample the neighbor tile. In our implementation, if the neighborhood is 100 pixels, then the center of the neighbor tile must be within 100 pixels of the anchor tile center both vertically and horizontally. The distant tile is sampled at random from outside this region.
\end{itemize}

Tile size should be chosen so that tiles are large enough to contain information at the scale needed for downstream tasks.
Neighborhood should be small enough that neighbor tiles will be semantically similar to the anchor tile, but large enough to capture intra-class (and potentially some inter-class) variability.
In practice, we find that plotting some example triplets as in Fig. \ref{fig:sampling} allowed us to find reasonable values for these parameters. Results across tile size and neighborhood on our land cover classification experiment are reported in Table \ref{table:hplr}.

Pseudocode for sampling a dataset of triplets is given in Algorithm \ref{alg:sampling}.
Note that no knowledge of actual geographical locations is needed, so Tile2Vec can be applied to any dataset without knowledge of the data collection procedure.

%%%%%%%%%%%%%%%%%%%%%%%%
%%%%%%%%%%%%%%%%%%%%%%%%

\subsection{Scalability} \label{scalability}

Like most deep learning algorithms, the Tile2Vec objective (Eq. \ref{objective}) allows for mini-batch training on large datasets.
More importantly, the use of the triplet loss allows the training dataset to grow with a \emph{power law} relationship relative to the size of the available remote sensing data.
Concretely, assume that for a given remote sensing dataset we have a sampling budget of $N$ triplets, imposed perhaps by storage constraints.
If we train using the straightforward approach of Eq. \ref{objective}, we will iterate over $N$ training examples in each epoch.
However, we notice that in most cases the area covered by our dataset is much larger than the area of a single neighborhood.
For any tile $t$, the likelihood that any particular $t'$ in the other $(N-1)$ tiles is in its neighborhood is extremely low.
Therefore, at training time we can match any $(t_a, t_n)$ pair with any of the $3N$ tiles in the dataset to massively increase the number of unique example triplets that the network sees from $\mathcal{O}(N)$ to $\mathcal{O}(N^2)$.

In practice, we find that combining Tile2Vec with this data augmentation scheme to create massive datasets results in an algorithm that is easy to train, robust to hyperparameter choices, and resistant to overfitting.
This point will be revisited in section \ref{training}.

\begin{figure*}[t]
\centering
  \includegraphics[width=1\linewidth]{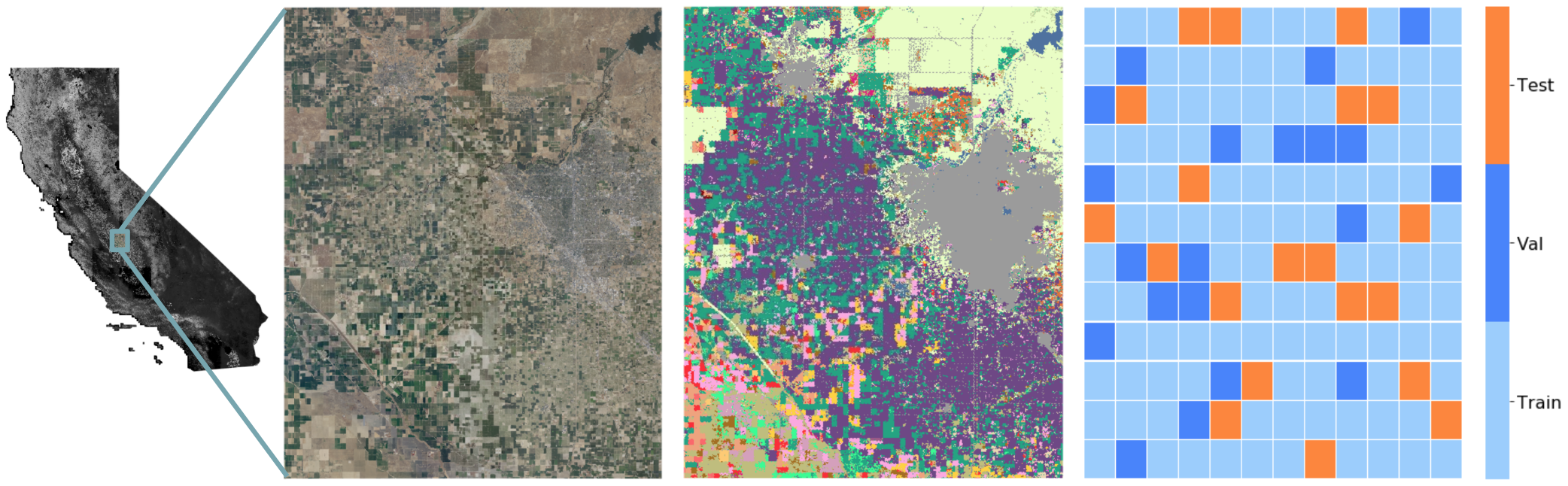}
  \caption{
  \textbf{Left:} Our NAIP aerial imagery of Central Valley covers 2500 km$^2$ around Fresno, California.
  \textbf{Center:} Land cover types in the region as labeled by the Cropland Data Layer (CDL, see Section \ref{dataset_cdl}) show a highly heterogeneous landscape; each color represents a different CDL class.
\textbf{Right:} For the land cover classification task, we split the dataset spatially into train, validation, and test sets. }
  \label{fig:naip}
\end{figure*}

\section{Datasets}

We evaluate Tile2Vec on several widely-used classes of remote sensing imagery, as well as a non-image dataset of country characteristics.
A brief overview of data organized by experiment is given here, with more detailed descriptions in Appendix \ref{appendix:datasets}.

%%%%%%%%%%%%%%%%%%%%%%%%

\subsection{Land cover classification}
\label{par:naip}

We first evaluate Tile2Vec on a land cover classification task --- predicting what is on the Earth's surface from remotely sensed imagery --- that uses the following two datasets:
The USDA's \textbf{National Agriculture Imagery Program (NAIP)} provides aerial imagery for public use that has four spectral bands --- red (R), green (G), blue (B), and infrared (N) --- at 0.6 m ground resolution.
We obtain an image of Central Valley, California near the city of Fresno for the year 2016 (Fig. \ref{fig:naip}), spanning latitudes $[36.45, 37.05]$ and longitudes $[-120.25, -119.65]$.
The \textbf{Cropland Data Layer (CDL)} is a raster geo-referenced land cover map collected by the USDA for the continental United States \cite{cdl}. \label{dataset_cdl}
Offered at 30 m resolution, it includes 132 class labels spanning crops, developed areas, forest, water, and more. In our NAIP dataset, we observe 66 CDL classes (Fig. \ref{fig:cdl}).
We use CDL as ground truth for evaluation by upsampling it to NAIP resolution.

%%%%%%%%%%%%%%%%%%%%%%%%

\subsection{Latent space interpolation and visual analogy}

We explore Tile2Vec embeddings by visualizing linearly interpolated tiles in the learned feature space and performing visual analogies on two datasets.
Tiles sampled from \textbf{NAIP} are used in the latent space interpolation evaluation.
The USGS and NASA's \textbf{Landsat 8 satellite} provide moderate-resolution (30 m) multi-spectral imagery on a 16-day collection cycle. Landsat datasets are public and widely used in agricultural, environmental, and other scientific applications. We generate median Landsat 8 composites containing 7 spectral bands over the urban and rural areas of three major US cities --- San Francisco, New York City, and Boston --- for a visual analogy evaluation.

%%%%%%%%%%%%%%%%%%%%%%%%

\subsection{Poverty prediction in Uganda}

Our next evaluation is a regression task: predicting local poverty levels from median \textbf{Landsat 7} composites of Uganda from 2009-2011 containing 5 spectral bands.
The World Bank's \textbf{Living Standards Measurement Study (LSMS)} surveys measure annual consumption expenditure at the household and village levels --- these measurements are the basis for determining international standards for extreme poverty. We use the Uganda 2011-12 survey as labels for the poverty prediction task described in \cite{jean2016combining}.

%%%%%%%%%%%%%%%%%%%%%%%%

\subsection{Worldwide country health index prediction}

Lastly, to demonstrate that Tile2Vec can be used with other high-dimensional vector data within a spatial context, we predict a subset of country characteristics from other country features.
The \textbf{CIA World Factbook} is an annual document compiled by the U.S. Central Intelligence Agency containing information on the governments, economies, energy systems, and societies of 267 world entities \cite{factbook2015world}.
We extract a dataset from the 2015 Factbook that contains 73 real-valued features (e.g. infant mortality rate, GDP per capita, crude oil production) for 242 countries.

%%% Copy in from notebook
\begin{table}
\footnotesize
\centering
\begin{tabular}{l c c c c c c}
\toprule
% & \multicolumn{6}{c}{Accuracy (\%)} \\
% \cmidrule(lr){2-7}
& \multicolumn{3}{c}{$n=1000$} & \multicolumn{3}{c}{$n=10000$}\\
\cmidrule(lr){2-4} \cmidrule(lr){5-7}
Unsupervised features & RF & LR & MLP & RF & LR & MLP\\
\midrule
Tile2Vec & \textbf{52.6} $\pm$ \textbf{1.1} & \textbf{53.7} $\pm$ \textbf{1.3} & \textbf{55.1} $\pm$ \textbf{1.2} & \textbf{56.9} $\pm$ \textbf{0.3} & \textbf{59.7} $\pm$ \textbf{0.3} & \textbf{58.4} $\pm$ \textbf{0.3} \\
Autoencoder & 49.1 $\pm$ 0.7 & 44.7 $\pm$ 1.0 & 52.0 $\pm$ 1.0 & 53.1 $\pm$ 0.2 & 55.6 $\pm$ 0.2 & 57.2 $\pm$ 0.4\\
Pre-trained ResNet-18 & 47.7 $\pm$ 0.6 & 48.4 $\pm$ 0.8 & 49.9 $\pm$ 1.7 & 50.6 $\pm$ 0.2 & 53.7 $\pm$ 0.2 & 54.4 $\pm$ 0.4\\
PCA & 46.9 $\pm$ 0.8 & 50.2 $\pm$ 0.4 & 43.6 $\pm$ 5.3 & 50.1 $\pm$ 0.3 & 51.1 $\pm$ 0.1 & 52.4 $\pm$ 0.3\\
ICA & 47.7 $\pm$ 0.6 & 50.1 $\pm$ 0.6 & 46.7 $\pm$ 3.1 & 50.4 $\pm$ 0.4 & 51.1 $\pm$ 0.1 & 52.5 $\pm$ 0.2\\
K-Means & 43.1 $\pm$ 0.8 & 49.4 $\pm$ 0.4 & 44.5 $\pm$ 3.9 & 45.6 $\pm$ 0.5 & 50.0 $\pm$ 0.1 & 50.5 $\pm$ 0.2\\

%%% NOT GENERATED AUTOMATICALLY
\bottomrule
\end{tabular}
\vspace{10pt}
\caption{Comparison of Tile2Vec features to unsupervised baselines on the CDL classification task in Section \ref{cdl}. Random forest (RF), logistic regression (LR), and multilayer perceptron (MLP) classifiers are trained over 10 trials of $n=1000$  and $n=10000$ randomly sampled labels, with mean accuracies and standard deviations reported. Baselines are described in detail in Section \ref{baselines}.}
\label{table:baselines}
\end{table}
\normalsize
%%%%%%%%%%%%%%%%%%%%%%%%%%%%%%%%%%%%%%%

\section{Experiments} \label{experiments}

%%%%%%%%%%%%%%%%%%%%%%%%

\subsection{Land cover classification using aerial imagery} \label{cdl}

We train Tile2Vec embeddings on 100k triplets sampled from the NAIP dataset from Central Valley, California.
The Tile2Vec CNN is a ResNet-18 architecture \cite{he2016deep} modified for $28\times28$ CIFAR-10 images (1) with an additional residual block to handle our larger input and (2) without the final classification layer.
Each of the 300k $50\times50$ NAIP tiles is labeled with the mode CDL land cover class and our evaluation metric is classification accuracy on this label.

%%%%%%%%%%%%%%%%%%%%%%%%

\subsubsection{Tile2Vec hyperparameter optimization}
\label{section:hp}

We tune the two main hyperparameters of Algorithm \ref{alg:sampling} on the NAIP dataset by searching over a grid of tile sizes and neighborhoods. In Table \ref{table:hplr}, accuracies on the CDL land cover classification task are reported across tile sizes of $[25,50,75,100]$ and neighborhoods of $[50,100,500,1000,\text{None}]$, where None indicates that both the neighbor and distant tiles are sampled from anywhere in the dataset.
Results suggest that, on this task and dataset, a neighborhood radius of 100 pixels strikes the ideal balance between sampling semantically similar tiles and capturing intra-class variability, though classification accuracy remains higher than the null model even when the neighborhood is increased to 1000 pixels. Accuracy also increases with tile size, which can be attributed to greater imbalance of labels at larger tile sizes (Appendix \ref{appendix:hp}) as well as greater available spatial context for classification.

Because CDL labels are at a resolution (30 m) equivalent to 50 NAIP pixels (0.6 m), we ultimately choose a tile size of 50 and neighborhood of 100 pixels for the land cover classification task. For consistency, subsequent experiments also use these hyperparameters and yield high performance. We suggest these hyperparameters be optimized anew for different datasets and tasks.

%%%%%%%%%%%%%%%%%%%%%%%%

\begin{figure}[t]
\centering
  \includegraphics[width=0.9\linewidth]{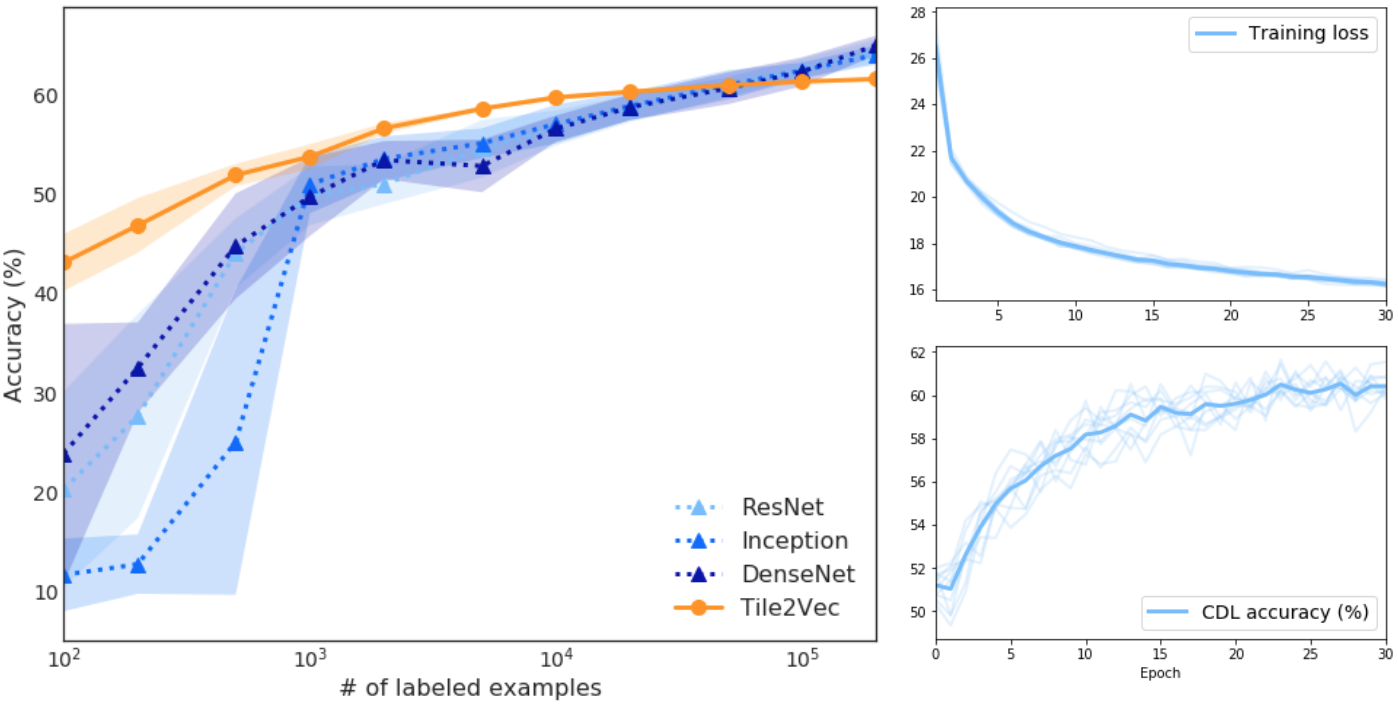}
  \caption{\textbf{Left:} Logistic regression on Tile2Vec unsupervised features outperforms supervised CNNs until 50k labeled examples.
  \textbf{Right:} The Tile2Vec triplet loss decreases steadily and performance on the downstream classification task tracks the loss closely.
  }
  \label{fig:supervised_loss}
\end{figure}

\subsubsection{Unsupervised learning baselines} \label{baselines}
We compare Tile2Vec to a number of unsupervised feature extraction methods.
We describe each baseline here, and provide additional training details in \ref{appendix:baselines}.

\vspace{-0.2cm}
\begin{itemize}[leftmargin=0.4cm]
\item \textbf{Autoencoder}: A convolutional autoencoder is trained on all 300k multi-spectral tiles, split 90\% training and 10\% validation. We train until the validation reconstruction error flattened; the encoder is then used to embed tiles into the feature space. The autoencoder achieves good reconstructions on the held-out test set (examples in \ref{appendix:autoencoder}).
\vspace{-0.1cm}
\item \textbf{Pre-trained ResNet-18}: A modified ResNet-18 was trained on resized CIFAR-10 images and used as a feature extractor. Since CIFAR-10 only has RGB channels, this approach only allows for use of the RGB bands of NAIP and illustrates the limitations of transferring models from natural images to remote sensing datasets.
\vspace{-0.1cm}
\item \textbf{PCA/ICA}: Each RGBN tile of shape $(50,50,4)$ is unraveled into a vector of length 10,000 and then PCA/ICA is used to compute the first 10 principal components for each tile.
\vspace{-0.1cm}
\item \textbf{K-means}: Tiles are clustered in pixel space using k-means with $k=10$, and each tile is represented as 10-dimensional vectors of distances to each cluster centroid.
\end{itemize}
\vspace{-0.1cm}
The features learned by Tile2Vec outperform other unsupervised features when used by random forest (RF), logistic regression (LR), and multilayer perceptron (MLP) classifiers trained on $n=1000$ or $n=10000$ labels (Table \ref{table:baselines}).
We also trained a DCGAN \cite{radford2015unsupervised} as a generative modeling approach to unsupervised feature learning.
Although we were able to generate reasonable samples, features learned by the discriminator performed poorly --- samples and results can be found in \ref{appendix:dcgan}.
Approaches based on variational autoencoders (VAEs) would also provide intriguing baselines, but we are unaware of existing models capable of capturing complex multi-spectral image distributions.

%%%%%%%%%%%%%%%%%%%%%%%%

\subsubsection{Supervised learning comparisons}
\label{supervised}

Surprisingly, our Tile2Vec features are also able to outperform fully-supervised CNNs trained directly on the classification task with large amounts of labeled data.
Fig. \ref{fig:supervised_loss} shows that applying logistic regression on Tile2Vec features beats several state-of-the-art supervised architectures \cite{he2016deep,szegedy2015going,huang2017densely} trained on as many as 50k CDL labels. 
We emphasize that the Tile2Vec CNN and the supervised ResNet share the same architecture, so logistic regression in Fig. \ref{fig:supervised_loss} is directly comparable to the classification layers of the supervised architectures.
Similar results for random forest and multi-layer perceptron classifiers can be found in Appendix \ref{appendix:supervised}.

%%%%%%%%%%%%%%%%%%%%%%%%

\begin{figure}[t]
\centering
  \includegraphics[width=0.8\linewidth]{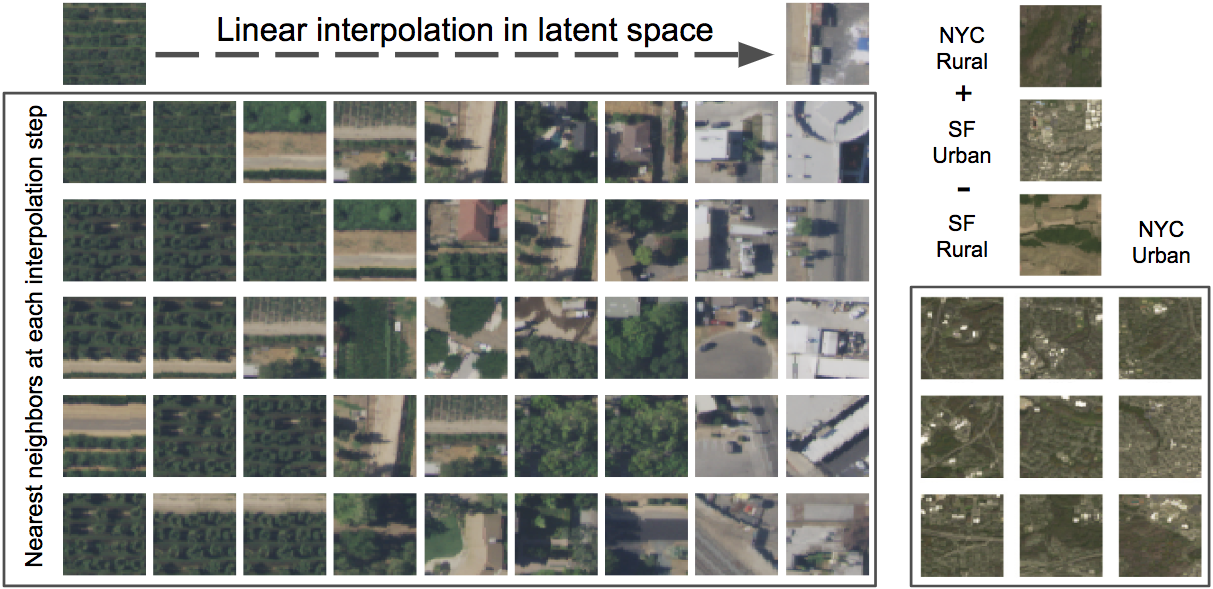}
  \setlength{\belowcaptionskip}{-3pt}
  \caption{\textbf{Left:} Linear interpolation in the latent space at equal intervals between representations of rural and urban images in the top row. Below, we show 5 nearest neighbors in the latent space to each interpolated vector. \textbf{Right:} Starting with a rural NYC embedding, we add urban SF and subtract rural SF to successfully discover urban NYC tiles. More visual analogies can be found in Fig. \ref{fig:analogies}.
  }
  \label{fig:latent}
\end{figure}

\subsubsection{Latent space interpolation}

We further explore the learned representations with a latent space interpolation experiment shown in Fig. \ref{fig:latent}.
Here, we start with the Tile2Vec embeddings of a field tile and an urban tile and linearly interpolate between the two.
At each point along the interpolation, we search for the five nearest neighbors in the latent space and display the corresponding tiles.
As we move through the semantically meaningful latent space, we recover tiles that are more and more developed.

%%%%%%%%%%%%%%%%%%%%%%%%

\subsubsection{Training details} \label{training}

Tile2Vec is easy to train and robust to the choice of hyperparameters.
We experimented with margins ranging from 0.1 to 100 and found little effect on accuracy.
Using a margin of 50, we trained Tile2Vec for 10 trials with different random initializations and show the results in Fig. \ref{fig:supervised_loss}.
The training loss is stable from epoch to epoch, consistently decreasing, and most importantly, a good proxy for unsupervised feature quality as measured by performance on the downstream task.

Tile2Vec learns good features even without regularizing the magnitudes of the learned embeddings (Eq. \ref{objective}).
However, by combining explicit regularization with the data augmentation scheme described in Section \ref{scalability}, we observe that Tile2Vec does not seem to overfit even when trained for many epochs.

%%%%%%%%%%%%%%%%%%%%%%%%

\subsection{Visual analogies across US cities}
\label{add_rs_exp}

To evaluate Tile2Vec qualitatively, we explore three major metropolitan areas of the United States: San Francisco, New York City, and Boston.
First, we train a Tile2Vec model on the San Francisco dataset only.
Then we use the trained model to embed tiles from all three cities.
As shown in Fig. \ref{fig:latent} and \ref{fig:analogies}, these learned representations allow us to perform arithmetic in the latent space, or \emph{visual analogies} \cite{radford2015unsupervised}.
By adding and subtracting vectors in the latent space, we can recover image tiles that are semantically what we would expect given the operations applied.

Here we use the full Landsat images with 7 spectral bands, demonstrating that Tile2Vec can be applied effectively to highly multi-spectral datasets.
Tile2Vec can also learn representations at multiple scales: each Landsat 8 (30 m resolution) tile covers 2.25 km$^2$, while the NAIP and DigitalGlobe tiles are 2500 times smaller.
Finally, Tile2Vec learns robust representations that allow for domain adaptation or transfer learning, as the three datasets have widely varying spectral distributions (Fig. \ref{fig:landsat}).

%%%%%%%%%%%%%%%%%%%%%%%%

\subsection{Poverty prediction from satellite imagery in Uganda}

Next, we apply Tile2Vec to predict annual consumption expenditures in Uganda from Landsat satellite imagery.
The previous state-of-the-art result used a transfer learning approach in which a CNN is trained to predict nighttime lights (a proxy for poverty) from daytime satellite images --- the features from this model are then used to predict consumption expenditures \cite{jean2016combining}. We use the same LSMS pre-processing pipeline and ridge regression evaluation (see \cite{jean2016combining} for details).
Evaluating over $10$ trials of 5-fold cross-validation, we report an average $r^2$ of $0.496 \pm 0.014$ compared to $r^2=0.41$ for the transfer learning approach --- this is achieved with publicly available daytime satellite imagery with much lower resolution than the proprietary images used in \cite{jean2016combining} (30 m vs. 2.4 m).

%%%%%%%%%%%%%%%%%%%%%%%%

%%% Copy in from notebook
\begin{table}
\footnotesize
\centering
\begin{tabular}{l c c c c}
\toprule
Features & d & kNN & RF & RR \\
\midrule
Tile2Vec & 10 & \textbf{77.5} $\pm$ \textbf{1.0} & \textbf{76.0} $\pm$ \textbf{1.3} & \textbf{69.6} $\pm$ \textbf{1.0} \\
Non-health & 60 & 62.8 $\pm$ 1.5 & 72.1 $\pm$ 1.6 & 68.7 $\pm$ 1.7 \\
Locations & 2 & 69.3 $\pm$ 1.0 & 67.7 $\pm$ 2.6 & 11.6 $\pm$ 1.5 \\

%%% NOT GENERATED AUTOMATICALLY
\bottomrule
\end{tabular}
\vspace{10pt}
\setlength{\belowcaptionskip}{-5pt}
\caption{Predicting health index using Tile2Vec features versus non-health features and locations (i.e. \{lat,lon\}). Here, $d$ is feature dimension, kNN is $k$-nearest neighbors, RF is random forest, and RR is ridge regression. Hyperparameters (e.g. $k$ and regularization strength) are tuned for each feature set. We report average $r^2$ and standard deviation for 10 trials of 3-fold cross-validation.}
\label{table:factbook}
\end{table}
%%%%%%%%%%%%%%%%%%%%%%%%%%%%%%%%%%%%%%%

\subsection{Generalizing to other spatial data: Predicting country health index from CIA Factbook}

To demonstrate that Tile2Vec can leverage spatial coherence for non-image datasets as well, we use 13 of the features in the CIA Factbook related to public health and compute a health index, then attempt to predict this health index from the remaining 60 features.
We train Tile2Vec by sampling triplets of countries and feeding the  feature vectors into a small fully-connected neural network with one hidden layer.
As shown in Table \ref{table:factbook}, the embeddings learned by Tile2Vec on this small spatial dataset ($N=242$) outperform both the original features and approaches that explicitly use spatial information.
Visualizations of the learned embeddings and full training details are included in \ref{appendix:factbook}.

\section{Related Work}

Our inspiration for using spatial context to learn representations originated from continuous word representations like Word2vec and GloVe \cite{mikolov2013distributed,mikolov2013efficient,pennington2014glove}.
In NLP, the distributional hypothesis can be summarized as ``a word is characterized by the company it keeps'' --- words that appear in the same context likely have similar semantics.
We apply this concept to remote sensing data, with multi-spectral image tiles as the atomic unit analogous to individual words in NLP, and geospatial neighborhoods as the ``company'' that these tiles keep.
A related, supervised version of this idea is the patch2vec algorithm \cite{fried2017patch2vec}, which its authors describe as learning ``globally consistent image patch representations''.
Working with natural images, they use a very similar triplet loss (first introduced in \cite{hoffer2015deep}), but sample their patches \emph{with supervision} from an annotated semantic segmentation dataset.

Unsupervised learning for visual data is an active area of research in machine learning today and thus impossible to summarize concisely, but we attempt a brief overview of the most relevant topics here.
The three main classes of deep generative models --- likelihood-based variational autoencoders (VAEs) \cite{kingma2013auto}, likelihood-free generative adversarial networks (GANs) \cite{goodfellow2014generative}, and various autoregressive models \cite{oord2016pixel,van2016conditional} --- attempt to learn the generating data distribution from training samples.
Other related lines of work use spatial or temporal context to learn high-level image representations.
Some strategies for using spatial context involve predicting the relative positions of patches sampled from within an image \cite{noroozi2016unsupervised,doersch2015unsupervised} or trying to fill in missing portions of an image (in-painting) \cite{pathak2016context}.
In videos, nearby frames can be used to learn temporal embeddings \cite{ramanathan2015learning}; other methods leveraging the temporal coherence and invariances of videos for feature learning have also been proposed \cite{misra2016shuffle,wang2015unsupervised}.

\section{Conclusion}

We demonstrate the efficacy of Tile2Vec as an unsupervised feature learning algorithm for spatially distributed data on tasks from land cover classification to poverty prediction. Our method can be applied to image datasets spanning moderate to high resolution, RGB or multi-spectral bands, and collected via aerial or satellite sensors, and even to non-image datasets. Tile2Vec outperforms other unsupervised feature extraction techniques on a difficult classification task --- surprisingly, it even outperforms supervised CNNs trained on 50k labeled examples.

In this paper, we focus on exploiting \emph{spatial} coherence, but many geospatial datasets also include sequences of data from the same locations collected over time.
Temporal patterns can be highly informative (e.g. seasonality, crop cycles), and we plan to explore this aspect in future work.
Remote sensing data have largely been unexplored by the machine learning community --- more research in these areas could result in enormous progress on many problems of global significance.

%%%%%%%%%%%%%%%%%%%%%%%%
%%%%%%%%%%%%%%%%%%%%%%%%

\newpage
% \printbibliography
\bibliographystyle{plain}
\bibliography{0_refs}

\begin{thebibliography}{10}

\bibitem{cdl}
{USDA National Agricultural Statistics Service Cropland Data Layer}. published
  crop-specific data layer [online]., 2016.

\bibitem{doersch2015unsupervised}
Carl Doersch, Abhinav Gupta, and Alexei~A Efros.
\newblock Unsupervised visual representation learning by context prediction.
\newblock In {\em Proceedings of the IEEE International Conference on Computer
  Vision}, pages 1422--1430, 2015.

\bibitem{factbook2015world}
CIA Factbook.
\newblock The world factbook; 2010.
\newblock {\em See also: http://www cia
  gov/library/publications/the-world-factbook, accessed January}, 30, 2015.

\bibitem{foody}
G.~M. Foody.
\newblock Remote sensing of tropical forest environments: Towards the
  monitoring of environmental resources for sustainable development.
\newblock {\em International Journal of Remote Sensing}, 24(20):4035--4046,
  2003.

\bibitem{fried2017patch2vec}
O~Fried, S~Avidan, and D~Cohen-Or.
\newblock Patch2vec: Globally consistent image patch representation.
\newblock In {\em Computer Graphics Forum}, volume~36, pages 183--194. Wiley
  Online Library, 2017.

\bibitem{goodfellow2014generative}
Ian Goodfellow, Jean Pouget-Abadie, Mehdi Mirza, Bing Xu, David Warde-Farley,
  Sherjil Ozair, Aaron Courville, and Yoshua Bengio.
\newblock Generative adversarial nets.
\newblock In {\em Advances in neural information processing systems}, pages
  2672--2680, 2014.

\bibitem{Gorelick:2017hd}
Noel Gorelick, Matt Hancher, Mike Dixon, Simon Ilyushchenko, David Thau, and
  Rebecca Moore.
\newblock {Google Earth Engine: Planetary-scale geospatial analysis for
  everyone}.
\newblock {\em Remote Sensing of Environment}, pages 1--10, aug 2017.

\bibitem{he2016deep}
Kaiming He, Xiangyu Zhang, Shaoqing Ren, and Jian Sun.
\newblock Deep residual learning for image recognition.
\newblock In {\em Proceedings of the IEEE conference on computer vision and
  pattern recognition}, pages 770--778, 2016.

\bibitem{hoffer2015deep}
Elad Hoffer and Nir Ailon.
\newblock Deep metric learning using triplet network.
\newblock In {\em International Workshop on Similarity-Based Pattern
  Recognition}, pages 84--92. Springer, 2015.

\bibitem{huang2017densely}
Gao Huang, Zhuang Liu, Kilian~Q Weinberger, and Laurens van~der Maaten.
\newblock Densely connected convolutional networks.
\newblock In {\em Proceedings of the IEEE conference on computer vision and
  pattern recognition}, volume~1, page~3, 2017.

\bibitem{jean2016combining}
Neal Jean, Marshall Burke, Michael Xie, W~Matthew Davis, David~B Lobell, and
  Stefano Ermon.
\newblock Combining satellite imagery and machine learning to predict poverty.
\newblock {\em Science}, 353(6301):790--794, 2016.

\bibitem{kingma2013auto}
Diederik~P Kingma and Max Welling.
\newblock Auto-encoding variational bayes.
\newblock {\em arXiv preprint arXiv:1312.6114}, 2013.

\bibitem{krizhevsky2012imagenet}
Alex Krizhevsky, Ilya Sutskever, and Geoffrey Hinton.
\newblock {ImageNet} classification with deep convolutional neural networks.
\newblock In {\em Advances in neural information processing systems}, pages
  1097--1105, 2012.

\bibitem{levin2010terrapattern}
G~Levin, D~Newbury, K~McDonald, I~Alvarado, A~Tiwari, and M~Zaheer.
\newblock Terrapattern: open-ended, visual query-by-example for satellite
  imagery using deep learning, 2010.

\bibitem{levy2014neural}
Omer Levy and Yoav Goldberg.
\newblock Neural word embedding as implicit matrix factorization.
\newblock In {\em Advances in neural information processing systems}, pages
  2177--2185, 2014.

\bibitem{Luck:2016bj}
W~L{\"u}ck and A~van Niekerk.
\newblock {Evaluation of a rule-based compositing technique for Landsat-5 TM
  and Landsat-7 ETM+ images}.
\newblock {\em International Journal of Applied Earth Observation and
  Geoinformation}, 47:1--14, May 2016.

\bibitem{madden2009manual}
M.~Madden, American~Society for Photogrammetry, and Remote Sensing.
\newblock {\em Manual of Geographic Information Systems}.
\newblock American Society for Photogrammetry and Remote Sensing, 2009.

\bibitem{mikolov2013efficient}
Tomas Mikolov, Kai Chen, Greg Corrado, and Jeffrey Dean.
\newblock Efficient estimation of word representations in vector space.
\newblock {\em arXiv preprint arXiv:1301.3781}, 2013.

\bibitem{mikolov2013distributed}
Tomas Mikolov, Ilya Sutskever, Kai Chen, Greg~S Corrado, and Jeff Dean.
\newblock Distributed representations of words and phrases and their
  compositionality.
\newblock In {\em Advances in neural information processing systems}, pages
  3111--3119, 2013.

\bibitem{misra2016shuffle}
Ishan Misra, C~Lawrence Zitnick, and Martial Hebert.
\newblock Shuffle and learn: unsupervised learning using temporal order
  verification.
\newblock In {\em European Conference on Computer Vision}, pages 527--544.
  Springer, 2016.

\bibitem{mulla}
David~J. Mulla.
\newblock Twenty five years of remote sensing in precision agriculture: Key
  advances and remaining knowledge gaps.
\newblock {\em Biosystems Engineering}, 114(4):358 -- 371, 2013.
\newblock Special Issue: Sensing Technologies for Sustainable Agriculture.

\bibitem{noroozi2016unsupervised}
Mehdi Noroozi and Paolo Favaro.
\newblock Unsupervised learning of visual representations by solving jigsaw
  puzzles.
\newblock In {\em European Conference on Computer Vision}, pages 69--84.
  Springer, 2016.

\bibitem{oord2016pixel}
Aaron van~den Oord, Nal Kalchbrenner, and Koray Kavukcuoglu.
\newblock Pixel recurrent neural networks.
\newblock {\em arXiv preprint arXiv:1601.06759}, 2016.

\bibitem{pathak2016context}
Deepak Pathak, Philipp Krahenbuhl, Jeff Donahue, Trevor Darrell, and Alexei~A
  Efros.
\newblock Context encoders: Feature learning by inpainting.
\newblock In {\em Proceedings of the IEEE Conference on Computer Vision and
  Pattern Recognition}, pages 2536--2544, 2016.

\bibitem{scikit-learn}
F.~Pedregosa, G.~Varoquaux, A.~Gramfort, V.~Michel, B.~Thirion, O.~Grisel,
  M.~Blondel, P.~Prettenhofer, R.~Weiss, V.~Dubourg, J.~Vanderplas, A.~Passos,
  D.~Cournapeau, M.~Brucher, M.~Perrot, and E.~Duchesnay.
\newblock Scikit-learn: Machine learning in {P}ython.
\newblock {\em Journal of Machine Learning Research}, 12:2825--2830, 2011.

\bibitem{pennington2014glove}
Jeffrey Pennington, Richard Socher, and Christopher Manning.
\newblock Glove: Global vectors for word representation.
\newblock In {\em Proceedings of the 2014 conference on empirical methods in
  natural language processing (EMNLP)}, pages 1532--1543, 2014.

\bibitem{radford2015unsupervised}
Alec Radford, Luke Metz, and Soumith Chintala.
\newblock Unsupervised representation learning with deep convolutional
  generative adversarial networks.
\newblock {\em arXiv preprint arXiv:1511.06434}, 2015.

\bibitem{ramanathan2015learning}
Vignesh Ramanathan, Kevin Tang, Greg Mori, and Li~Fei-Fei.
\newblock Learning temporal embeddings for complex video analysis.
\newblock In {\em Proceedings of the IEEE International Conference on Computer
  Vision}, pages 4471--4479, 2015.

\bibitem{Roy:2014fla}
D~P Roy, M~A Wulder, T~R Loveland, C~E Woodcock, R~G Allen, M~C Anderson,
  D~Helder, J~R Irons, D~M Johnson, R~Kennedy, T~A Scambos, C~B Schaaf, J~R
  Schott, Y~Sheng, E~F Vermote, A~S Belward, R~Bindschadler, W~B Cohen, F~Gao,
  J~D Hipple, P~Hostert, J~Huntington, C~O Justice, A~Kilic, V~Kovalskyy, Z~P
  Lee, L~Lymburner, J~G Masek, J~McCorkel, Y~Shuai, R~Trezza, J~Vogelmann, R~H
  Wynne, and Z~Zhu.
\newblock {Landsat-8: Science and product vision for terrestrial global change
  research}.
\newblock {\em Remote Sensing of Environment}, 145(C):154--172, April 2014.

\bibitem{szegedy2015going}
Christian Szegedy, Wei Liu, Yangqing Jia, Pierre Sermanet, Scott Reed, Dragomir
  Anguelov, Dumitru Erhan, Vincent Vanhoucke, Andrew Rabinovich, et~al.
\newblock Going deeper with convolutions.
\newblock Cvpr, 2015.

\bibitem{van2016conditional}
Aaron van~den Oord, Nal Kalchbrenner, Lasse Espeholt, Oriol Vinyals, Alex
  Graves, et~al.
\newblock Conditional image generation with pixelcnn decoders.
\newblock In {\em Advances in Neural Information Processing Systems}, pages
  4790--4798, 2016.

\bibitem{tsne}
L.J.P. van~der Maaten and Geoffrey Hinton.
\newblock Visualizing high-dimensional data using {t-SNE}.
\newblock {\em Journal of Machine Learning Research}, 9:2579--2605, 2008.

\bibitem{wang2015unsupervised}
Xiaolong Wang and Abhinav Gupta.
\newblock Unsupervised learning of visual representations using videos.
\newblock {\em arXiv preprint arXiv:1505.00687}, 2015.

\bibitem{Whitcraft:2015co}
Alyssa~K Whitcraft, Eric~F Vermote, Inbal Becker-Reshef, and Christopher~O
  Justice.
\newblock {Cloud cover throughout the agricultural growing season: Impacts on
  passive optical earth observations}.
\newblock {\em Remote Sensing of Environment}, 156:438--447, January 2015.

\end{thebibliography}

\newpage
\normalsize
\appendix

\setcounter{figure}{0}
\renewcommand\thefigure{\thesection\arabic{figure}}
\renewcommand{\thetable}{\thesection\arabic{table}}

%%%%%%%%%%%%%%%%%%%%%%%%

\section{Appendix}

\subsection{Baselines} \label{appendix:baselines}

\subsubsection{Autoencoder} \label{appendix:autoencoder}

\begin{figure}[h]
\centering
  \includegraphics[width=0.7\linewidth]{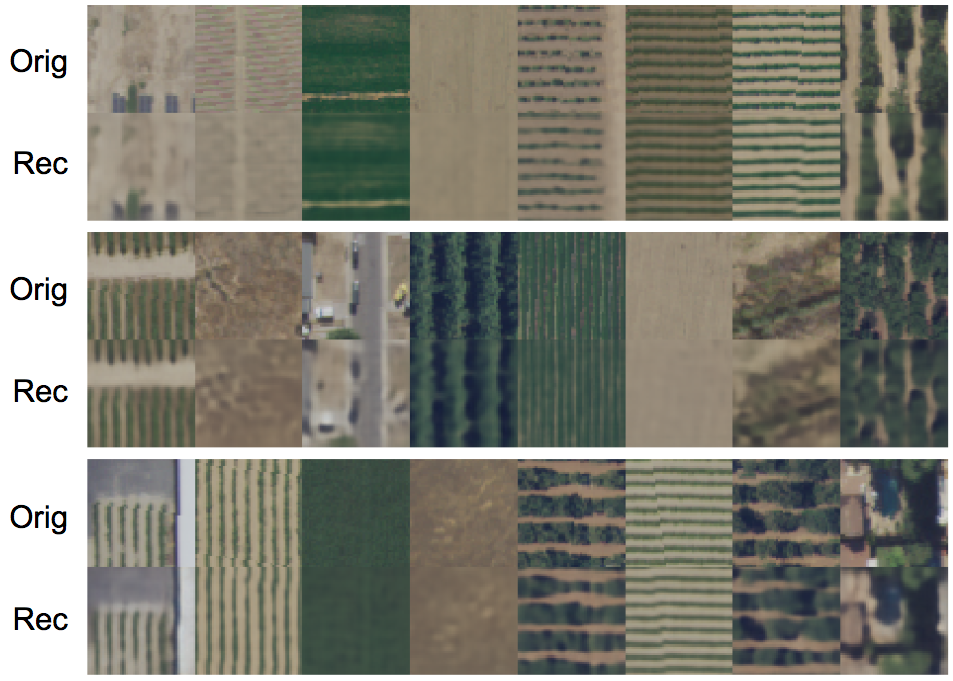}
  \caption{Reconstruction examples from convolutional autoencoder trained on NAIP dataset. Top rows contains original image tiles, bottom rows contains reconstructions --- all examples shown are from hold-out validation set that is not seen during autoencoder training.
  }
  \label{fig:autoencoder}
\end{figure}

The convolutional autoencoder architecture has 3 convolutional layers and 2 linear layers in the encoder and 1 linear layer and 3 deconvolutional layers in the decoder; it is trained with batch size 100.
Reconstructed samples can be seen in Fig. \ref{fig:autoencoder} --- the autoencoder is able to learn good reconstructions even for hold-out examples are that unseen at training time.

\subsubsection{DCGAN} \label{appendix:dcgan}

\begin{figure}[h]
\centering
  \includegraphics[width=0.6\linewidth]{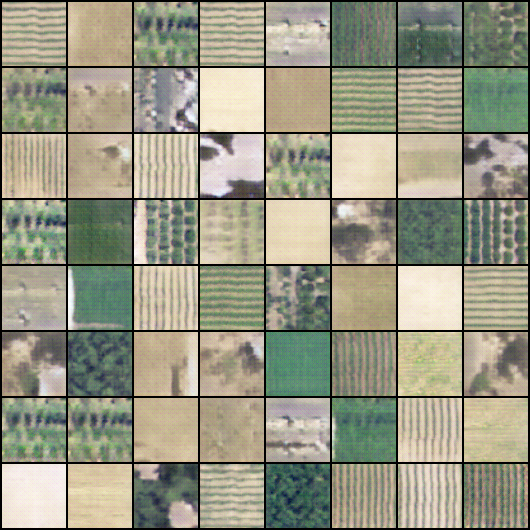}
  \caption{Visualized RGB bands of $50\times 50 \times 4$ DCGAN generated samples.}
  \label{fig:dcgan}
\end{figure}

We train a slightly modified DCGAN \cite{radford2015unsupervised} on 300k $50 \times 50 \times 4$ NAIP tiles.
Although the DCGAN was able to generate reasonable data samples (RGB channels visualized in Fig. \ref{fig:dcgan}), we were unable to achieve classification accuracies higher than 25\% using the features learned by the trained discriminator.

\subsubsection{PCA, ICA, K-means}
\label{appendix:baseline_other}

Non-deep learning feature extraction approaches such as PCA, ICA, and k-means are fit on randomly sampled subsets of 10,000 points for tractability.

\subsubsection{Supervised methods}
\label{appendix:supervised}

\begin{figure}[t]
\centering
  \includegraphics[width=0.9\linewidth]{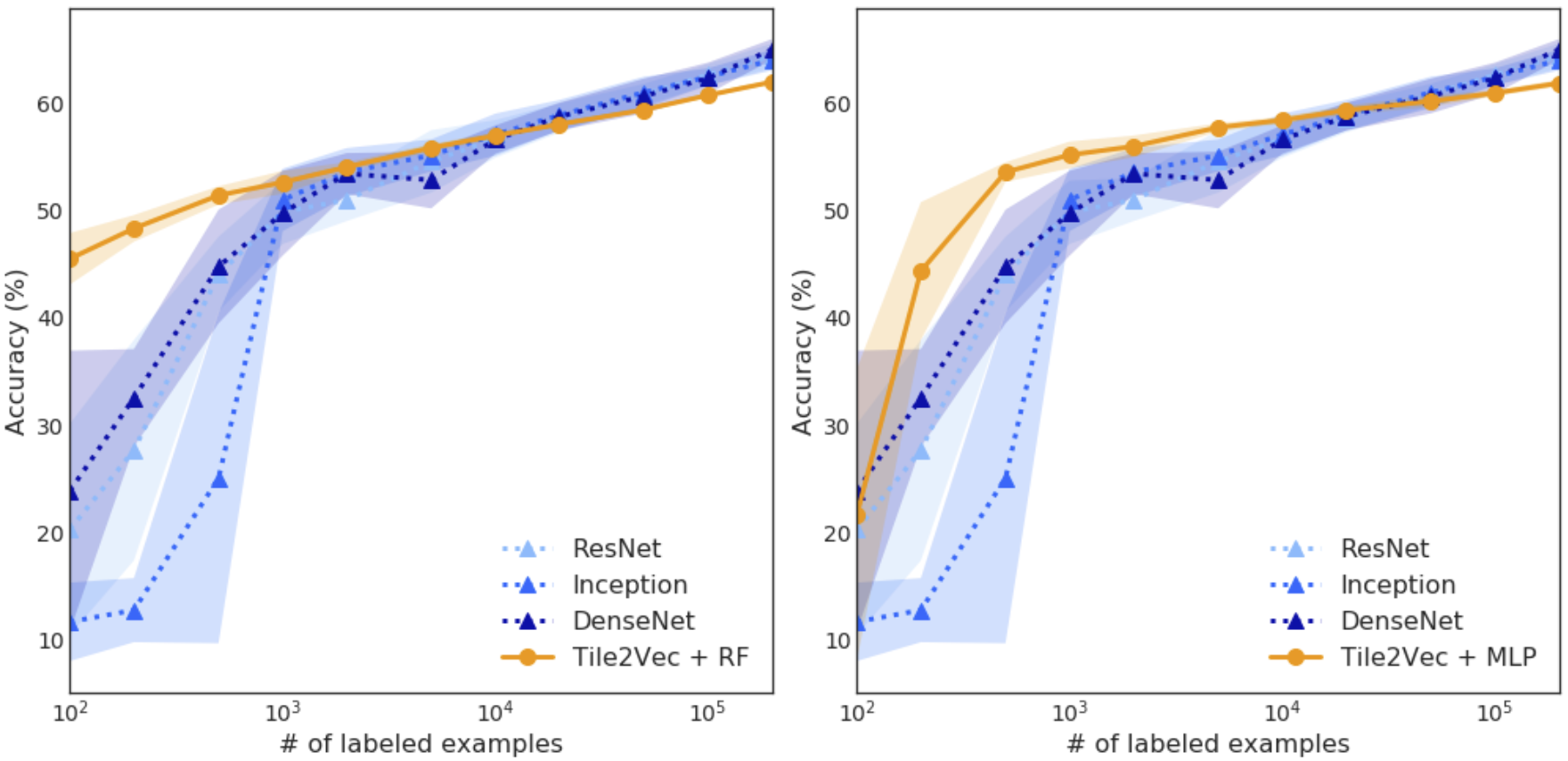}
  \caption{\textbf{Left:} Random forest (RF) on Tile2Vec unsupervised features compared to supervised baselines. \textbf{Right:} Multilayer perceptron (MLP) on Tile2Vec unsupervised features compared to supervised baselines.
  }
  \label{fig:supervised_rfmlp}
\end{figure}

As mentioned in Section \ref{supervised}, we train several state-of-the-art supervised architectures \cite{he2016deep,szegedy2015going,huang2017densely} on increasing amounts of data and find that Tile2Vec paired with a classifier (logistic regression, random forest, or multilayer perceptron) performs better for dataset sizes up to 50k. 
Results for random forest and multilayer perceptron classifiers are shown in Fig. \ref{fig:supervised_rfmlp}.

%%%%%%%%%%%%%%%%%%%%%%%%

\newpage
\subsection{Predicting poverty in Uganda}

\begin{figure}[t]
\centering
  \includegraphics[width=0.8\linewidth]{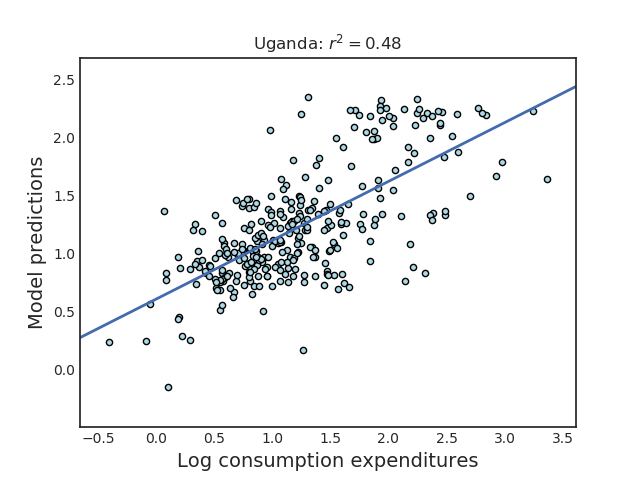}
  \caption{Ridge regression model fit to average Tile2Vec embeddings of Landsat 7 tiles. Previous best reported $r^2$ is $0.41$.}
  \label{fig:poverty}
\end{figure}

Here we expand on using Tile2Vec to predict annual consumption expenditure in Uganda from satellite imagery. For more information on Landsat and the imagery acquisition process, see Appendix \ref{appendix:landsat}.

We sample following Algorithm \ref{alg:sampling}. Each sampled tile spans $50 \times 50$ pixels (2.25 km$^2$). We train Tile2Vec embeddings on $100k$ such triplets for $50$ epochs using a margin of $50$, $0.01$ l2 regularization, and other details described in Appendix \ref{appendix:details}. 

To measure how well these embeddings predict poverty, we use the World Bank Living Standards Measurement Surveys (LSMS). This survey, conducted in Uganda in 2011-12, samples clusters throughout the country based on population and then randomly surveys households within each cluster. Per capita expenditures within each cluster are averaged across households. The final dataset includes $315$ clusters.

The best known result on Uganda LSMS data uses a transfer learning approach in which a CNN is trained to predict night lights (an indicator of poverty) from daytime satellite images --- the features from this model are then used to predict consumption expenditures. To compare our results to this established baseline, we use the same LSMS preprocessing pipeline as Jean \emph{et al.} \cite{jean2016combining}, as well as their evaluation method of fitting a ridge regression model to CNN feature embeddings. 

For each cluster, we extract a median composite through Google Earth Engine of roughly $75 \times 75$ pixels (5 km$^2$) centered at its location. We randomly sample $10$ tiles from this patch and average their Tile2Vec embeddings. These embeddings are then input to a ridge regression to predict log consumption expenditures. We compute average $r^2$ across five cross-validation folds.

After training for $50$ epochs and evaluating over $10$ 5-fold cross validation trials, we achieve $r^2$ of $0.496 \pm 0.014$. In comparison, the reported $r^2$ for Uganda using the transfer learning approach is $0.41$. In Fig. \ref{fig:poverty}, we show predicted vs. true values from one of the cross validation trials.

%%%%%%%%%%%%%%%%%%%%%%%%

\subsection{Worldwide health index prediction using CIA Factbook} \label{appendix:factbook}

\begin{figure}[H]
\centering
  \includegraphics[width=1\linewidth]{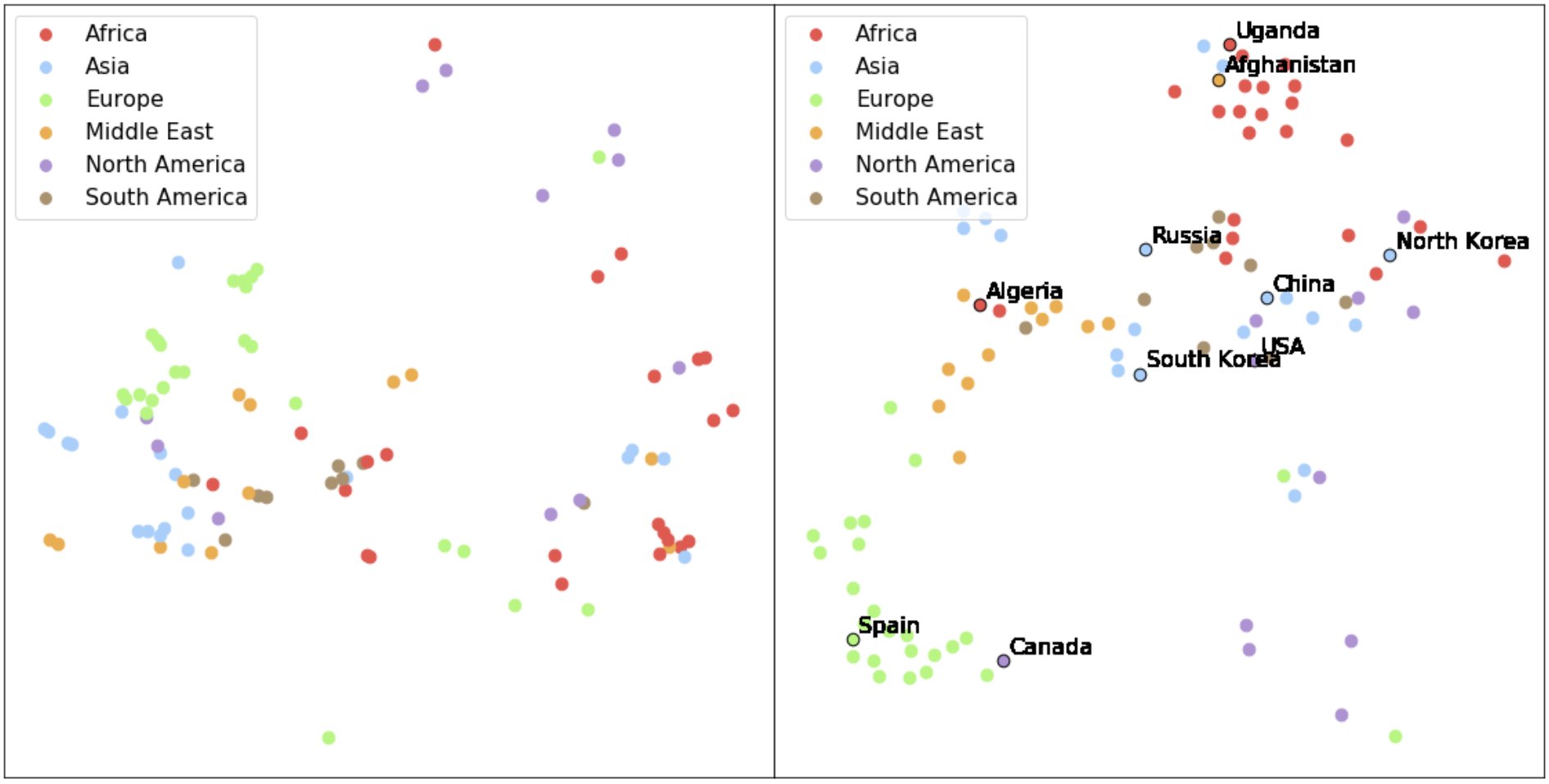}
  \caption{\textbf{Left:} The 60 non-health country features visualized using t-SNE. Spatial relationships are preserved for some clusters, but not for others. \textbf{Right:} The 10-dimensional Tile2Vec embeddings visualized using t-SNE. The latent space now respects both spatial and characteristic similarities. Several countries are annotated to highlight interesting relationships (e.g., North Korea and South Korea embedded far apart even though they are spatial neighbors; USA, South Korea, and China embedded close together though they are geographically separated).
  }
  \label{fig:factbook}
\end{figure}

The CIA Factbook contains 73 features spanning economic, energy, social, and other characteristics of countries around the world. We use 13 of the features related to public health and compute a health index, then attempt to predict this health index from the remaining 60 features. Fig. \ref{fig:factbook} shows the original 60-dimensional feature vectors as well as the 10-dimensional learned Tile2Vec embeddings projected down to two dimensions using t-SNE \cite{tsne}. While there is some geographic grouping of countries in projecting down the original features, the Tile2Vec embeddings appear to capture both geographic proximity and socioeconomic similarity.

%%%%%%%%%%%%%%%%%%%%%%%%

\subsection{Additional experiments} \label{appendix:additional}

\subsubsection{Visualizing NAIP embeddings}

\begin{figure}[h]
\centering
  \includegraphics[width=1\linewidth]{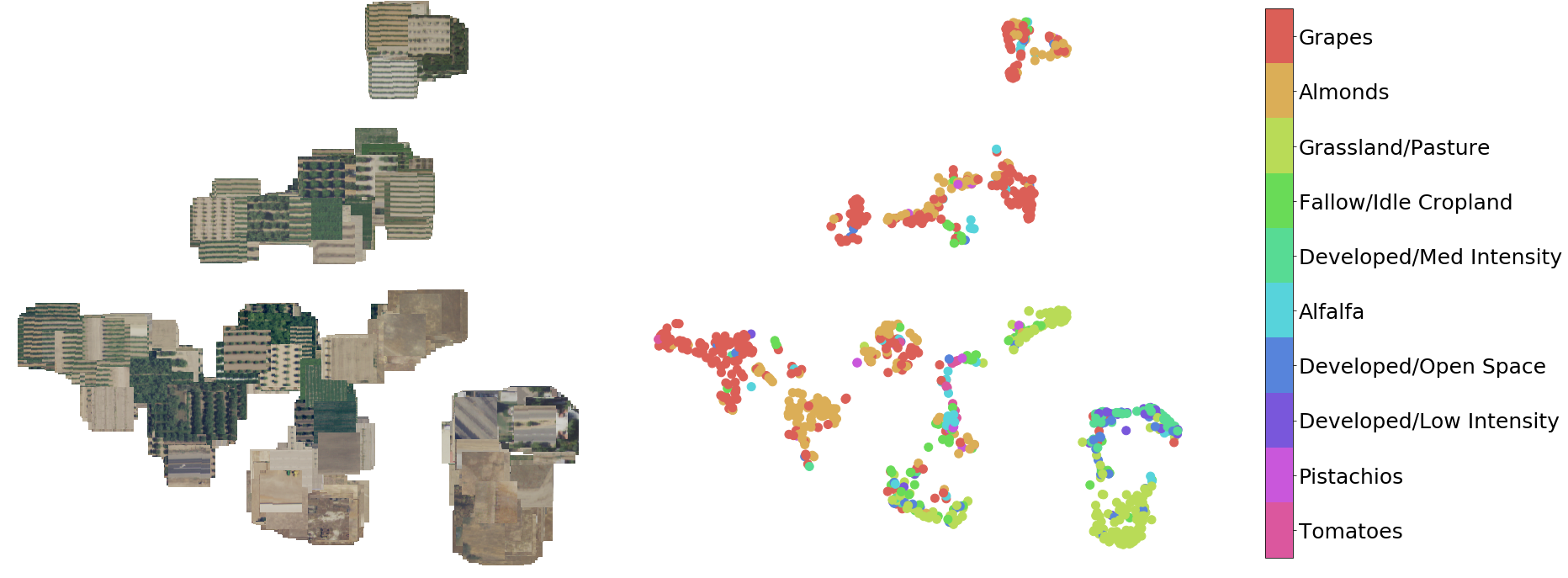}
  \caption{\textbf{Left:} NAIP image tiles visualized with t-SNE in the Tile2Vec learned embedding space. \textbf{Right:} The same learned embeddings for tiles represented as colored points where the colors correspond to the CDL label for each tile. Only the top 10 most common CDL classes are shown for clarity.}
  \label{fig:embeddings}
\end{figure}

Fig. \ref{fig:embeddings} shows that the learned Tile2Vec embedding clusters tiles from the same CDL class together in the latent space.
Note that some classes have multiple clusters, which can be understood by considering the high intra-class variability seen in Fig. \ref{fig:cdl}.
Some crops may look similar to other crops depending on when in the growing season the images are taken, how old the plants are, and many other reasons.
For the CDL classification task, we want to learn a representation that allows for clustering different crop types; for other potential downstream tasks, we may prefer a different clustering.
Being unsupervised, Tile2Vec is agnostic to the downstream application and learns a representation that obeys the distributional semantics of the image dataset.

\subsubsection{Visual analogies across US cities}

\begin{figure}[t]
\centering
  \includegraphics[width=0.7\linewidth]{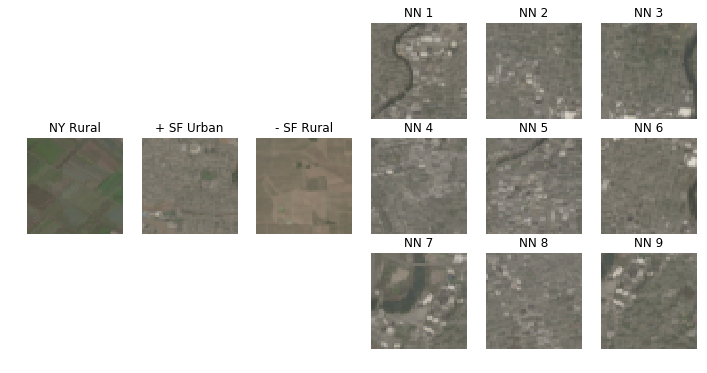}
  \includegraphics[width=0.7\linewidth]{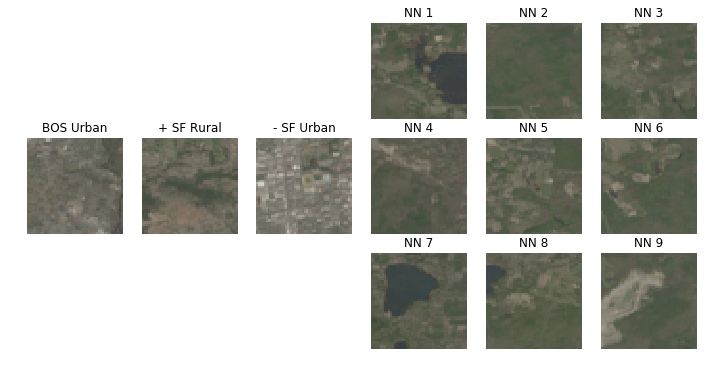}
  \caption{Additional visual analogy examples for Landsat SF, NYC, and BOS.}
  \label{fig:analogies}
\end{figure}

Recall from Section \ref{add_rs_exp} that we use the embeddings from three major metropolitan areas of the United States --- San Francisco, New York City, and Boston --- to perform analogies via arithmetic in our embedding space.
We show additional examples of analogies in Fig. \ref{fig:analogies}. Starting with a New York rural tile embedding, we add a San Francisco urban embedding and subtract a San Francisco rural embedding. The nearest neighbors of the resulting embedding belong to urban tiles in the New York dataset.

\begin{figure}[t]
\centering
  \includegraphics[width=\linewidth]{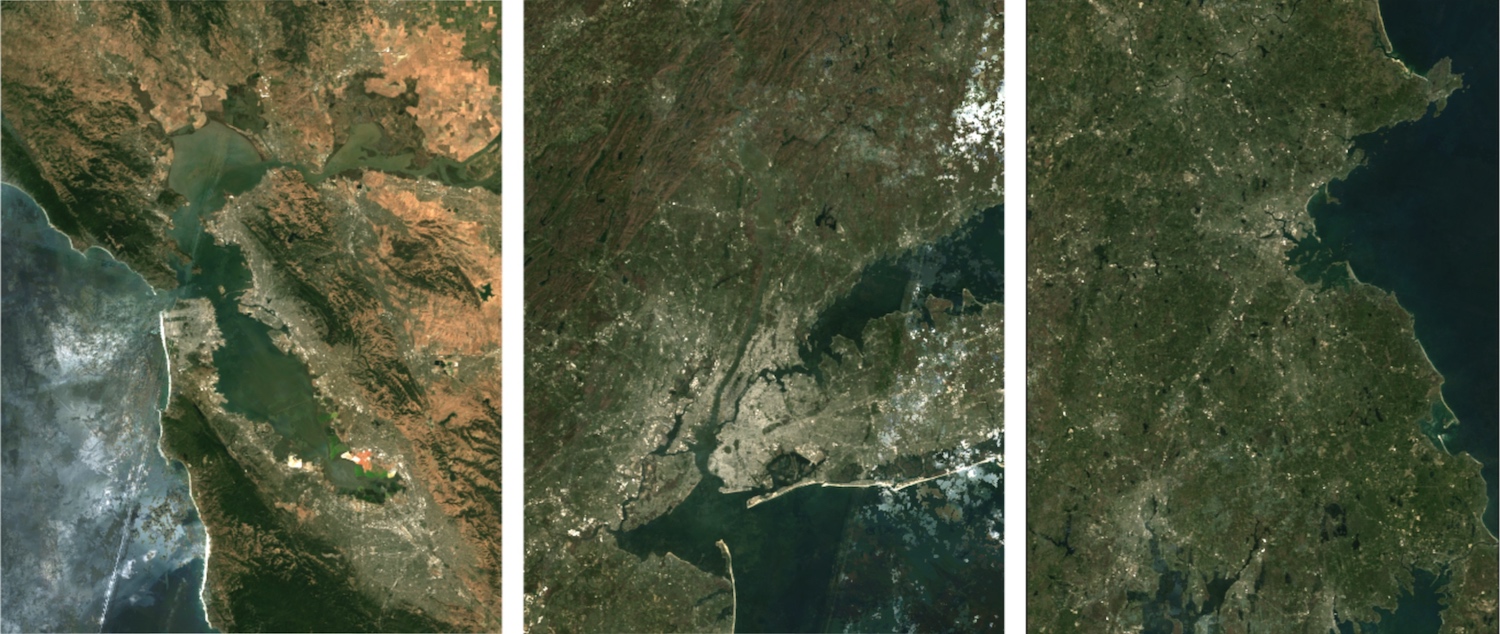}
  \caption{Landsat 8 median composites of three major US cities downloaded through Google Earth Engine: (Left) San Francisco, (Center) New York City, and (Right) Boston.}
  \label{fig:landsat}
\end{figure}

We highlight that the Tile2Vec CNN is trained only on Landsat imagery from San Francisco and generalizes to two cities on the east coast of the US with quite different spectral signatures (Fig. \ref{fig:landsat}).

\subsubsection{Visual query by example in Addis Ababa, Ethiopia}

We train a Tile2Vec model on high-resolution DigitalGlobe (see \ref{appendix:dg}) image tiles from Addis Ababa, Ethiopia to explore performance on (1) multiple modes of imaging (satellite vs. aerial) and (2) very different landscapes (sub-Saharan Africa vs. Central Valley California).
We also want to test the model in the developing world, where there is very little labeled data --- unsupervised methods for remote sensing could enable monitoring of infrastructure development and other humanitarian goals.

\begin{figure}[t]
\centering
  \includegraphics[width=0.65\linewidth]{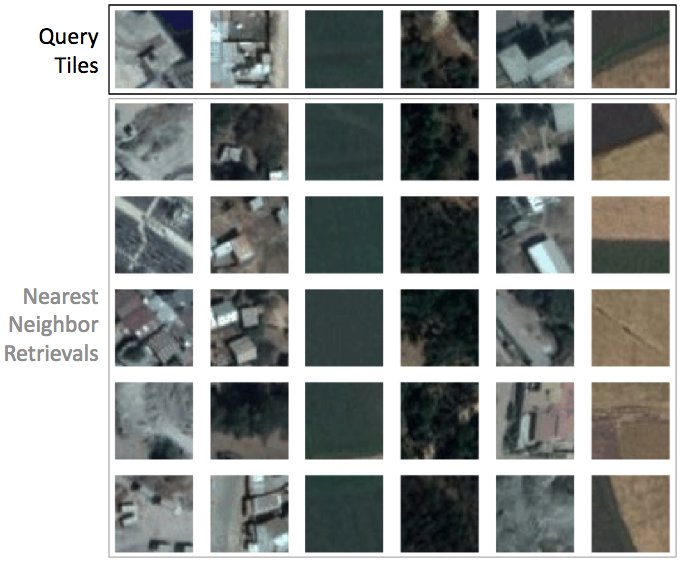}
  \caption{Visual query by example in Addis Ababa, Ethiopia. The top tile in each column is the query; its five nearest neighbors in the embedding space fill the rest of the column.}
  \label{fig:visualquery}
\end{figure}

In Fig. \ref{fig:visualquery}, we show that the latent representations learned by Tile2Vec allow us to do visual query by example: given a starting location, we can search for neighbors in the latent space to find other similar locations.
This type of visual query has previously been explored in the supervised setting \cite{levin2010terrapattern}, and has applications from military reconnaissance to disaster recovery. 

%%%%%%%%%%%%%%%%%%%%%%%%

\newpage
\subsection{Tile2Vec Hyperparameter Tuning}
\label{appendix:hp}

Across classifiers, we observe a neighborhood radius of 100 pixels to be optimal on the NAIP dataset for the task of land cover classification. Neighborhoods that are too large result in noisier triplets where the anchor and neighbor tile are not in semantically similar classes, while neighborhoods that are too small capture less intra-class variability. Larger tiles also perform better due to greater class imbalance and greater available spatial context.
Note that results in each row are directly comparable (i.e., the evaluation metric is exactly the same), while results in each column are not (i.e., evaluation task depends on tile size); thus, we italicize the best performing hyperparameters in each row and bold the overall best hyperparameter combination.

Results for a logistic regression classifier are shown in Table \ref{table:hplr}, random forest in Table \ref{table:hprf}, and multi-layer perceptron in Table \ref{table:hpmlp}. We show that the choice of hyperparameters is largely agnostic to the downstream classifier.
Also of note is the model performance with a neighborhood of None, which refers to sampling a neighbor tile from anywhere in the dataset and a distant tile from anywhere as well. We would expect no signal in this training strategy; indeed, the accuracies in the None column correspond to the frequency of the majority class in our test set at various tile sizes (Fig. \ref{fig:testsetdist}).

\begin{table}[h]
\footnotesize
\centering
\begin{tabular}{c c c c c c}
\toprule
& \multicolumn{5}{c}{Neighborhood radius} \\
\cmidrule(lr){2-6}
Tile size & 50 & 100 & 500 & 1000 & None\\
\midrule
25 & \textit{57.9} $\pm$ \textit{0.3} & 57.8 $\pm$ 0.4 & 53.4 $\pm$ 0.2 & 50.7 $\pm$ 0.2 & 34.1 $\pm$ 0.0\\
50 & 58.6 $\pm$ 0.3 & \textit{59.7} $\pm$ \textit{0.3} & 55.8 $\pm$ 0.2 & 49.7 $\pm$ 0.2 & 34.3 $\pm$ 0.0\\
75 & 58.9 $\pm$ 0.5 & \textit{59.8} $\pm$ \textit{0.4} & 58.2 $\pm$ 0.4 & 55.0 $\pm$ 0.3 & 34.5 $\pm$ 0.0\\
100 & 60.6 $\pm$ 0.1 & \textbf{\textit{61.8}} $\pm$ \textbf{\textit{0.4}} & 58.8 $\pm$ 0.4 & 55.1 $\pm$ 0.4 & 35.2 $\pm$ 0.0\\
\bottomrule
\end{tabular}
\vspace{10pt}
\caption{Logistic regression accuracies on land cover classification across hyperparameters tile size and neighborhood radius.}
\label{table:hplr}
\end{table}

\begin{table}[h]
\footnotesize
\centering
\begin{tabular}{c c c c c c}
\toprule
& \multicolumn{5}{c}{Neighborhood radius} \\
\cmidrule(lr){2-6}
Tile size & 50 & 100 & 500 & 1000 & None\\
\midrule
25 & \textit{55.3} $\pm$ \textit{0.3} & 54.7 $\pm$ 0.4 & 48.3 $\pm$ 0.3 & 46.3 $\pm$ 0.3 & 34.1 $\pm$ 0.0\\
50 & 56.7 $\pm$ 0.4 & \textit{56.9} $\pm$ \textit{0.3} & 50.4 $\pm$ 0.4 & 46.5 $\pm$ 0.4 & 34.3 $\pm$ 0.0\\
75 & 56.6 $\pm$ 0.4 & \textit{57.5} $\pm$ \textit{0.5} & 52.7 $\pm$ 0.4 & 49.7 $\pm$ 0.4 & 34.5 $\pm$ 0.0\\
100 & 58.6 $\pm$ 0.4 & \textbf{\textit{59.5}} $\pm$ \textbf{\textit{0.3}} & 53.1 $\pm$ 0.3 & 50.3 $\pm$ 0.4 & 35.2 $\pm$ 0.0\\
\bottomrule
\end{tabular}
\vspace{10pt}
\caption{Random forest classifier accuracies on land cover classification across hyperparameters tile size and neighborhood radius.}
\label{table:hprf}
\end{table}

\begin{table}[h]
\footnotesize
\centering
\begin{tabular}{c c c c c c}
\toprule
& \multicolumn{5}{c}{Neighborhood radius} \\
\cmidrule(lr){2-6}
Tile size & 50 & 100 & 500 & 1000 & None\\
\midrule
25 & \textit{57.2} $\pm$ \textit{0.4} & 56.6 $\pm$ 0.5 & 49.4 $\pm$ 0.7 & 46.7 $\pm$ 0.6 & 34.1 $\pm$ 0.0\\
50 & \textit{58.6} $\pm$ \textit{0.4} & 58.1 $\pm$ 0.5 & 50.4 $\pm$ 1.0 & 44.8 $\pm$ 0.6 & 34.3 $\pm$ 0.0\\
75 & 58.5 $\pm$ 0.5 & \textit{58.6} $\pm$ \textit{0.5} & 53.7 $\pm$ 1.4 & 47.7 $\pm$ 0.9 & 34.5 $\pm$ 0.0\\
100 & 60.7 $\pm$ 0.6 & \textbf{\textit{61.2}} $\pm$ \textbf{\textit{0.6}} & 52.0 $\pm$ 1.5 & 49.6 $\pm$ 0.9 & 35.2 $\pm$ 0.0\\
\bottomrule
\end{tabular}
\vspace{10pt}
\caption{Multi-layer perceptron classifier accuracies on land cover classification across hyperparameters tile size and neighborhood radius.}
\label{table:hpmlp}
\end{table}

%%%%%%%%%%%%%%%%%%%%%%%%

\newpage
\subsection{Datasets} \label{appendix:datasets}

\subsubsection{National Agriculture Imagery Program (NAIP)}

We obtain a large static aerial image from the National Agriculture Imagery Program (NAIP) of Central Valley, California near the city of Fresno for the year 2016 (Fig. \ref{fig:naip}).
The image spans latitudes $[36.45, 37.05]$ and longitudes $[-120.25, -119.65]$.
The study area contains a mixture of urban, suburban, agricultural, and other land use types; it was chosen for this diversity of land cover, which makes for a difficult classification task.

The NAIP imagery consists of four spectral bands --- red (R), green (G), blue (B), and infrared (N) --- at 0.6 m ground resolution.
Images of the study area amount to a 50 GB dataset containing over 12 billion multi-spectral pixels, and were exported piece-wise using Google Earth Engine \citep{Gorelick:2017hd}.
NAIP is an ideal remote sensing dataset in many ways: it is publicly accessible, has only cloud-free images, and its 0.6 m resolution allows a CNN to learn features (individual plants, small buildings, etc.) helpful for distinguishing land cover types that are only visible at high resolution.
%However, NAIP images are only collected every 3 to 5 years and in selected locations; therefore, it is not suitable for large-scale, continuous monitoring of land cover and land use change.

\subsubsection{Cropland Data Layer (CDL)}

\begin{figure}[t]
\centering
  \includegraphics[width=1\linewidth]{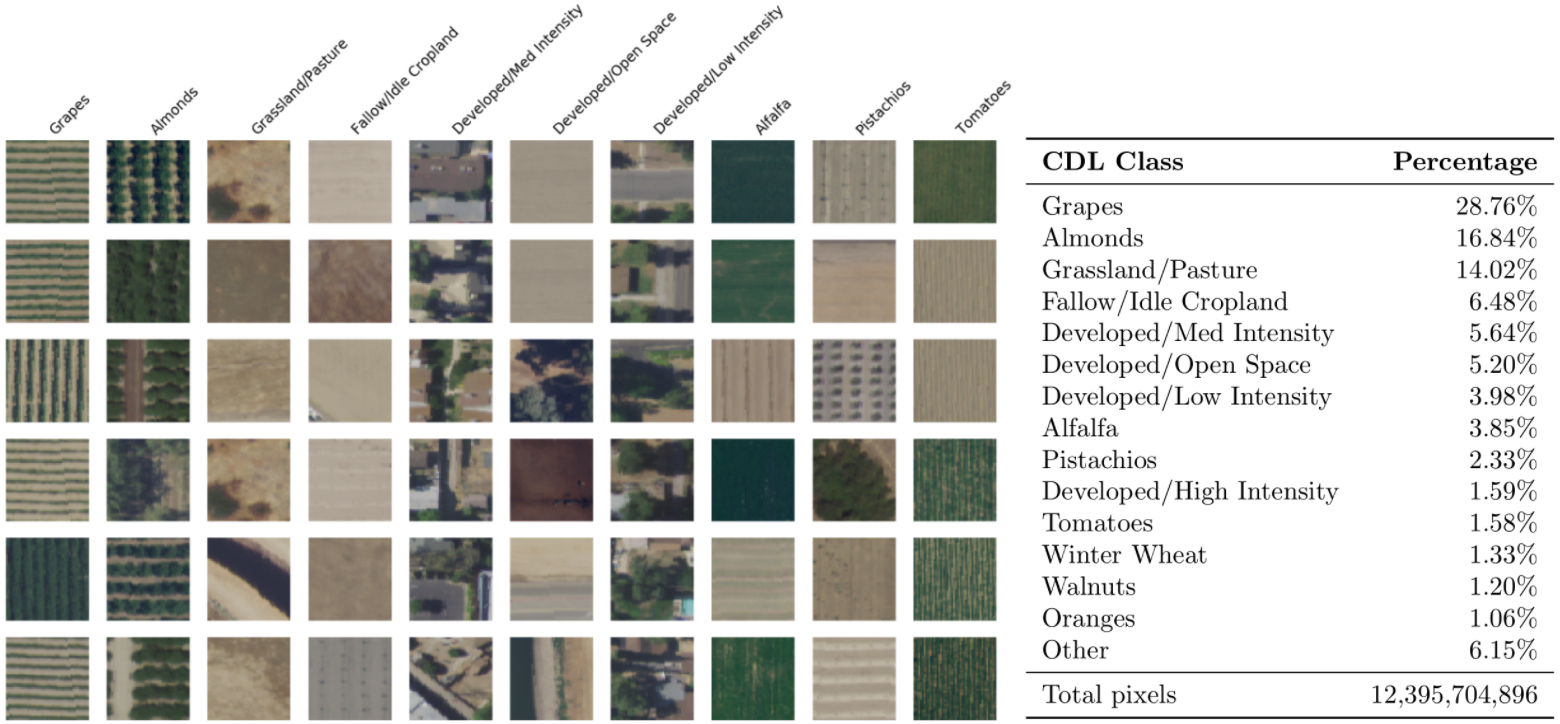}
  \caption{Six random examples of 50-by-50 pixel (30-by-30 m) tiles sampled from NAIP imagery. The top 10 classes in CDL (Section \ref{dataset_cdl}) are shown, with grapes being the most common. We label a tile with a CDL class if more than 80\% of pixels in the tile are in said class. There are 66 CDL classes in our entire dataset; the top 10 classes account for 88\% of tiles. Note the high within-class variability of tiles relative to between-class variability, which makes CDL classification a difficult task.}
  \label{fig:cdl}
\end{figure}

\begin{figure}[t]
\centering
  \includegraphics[width=1\linewidth]{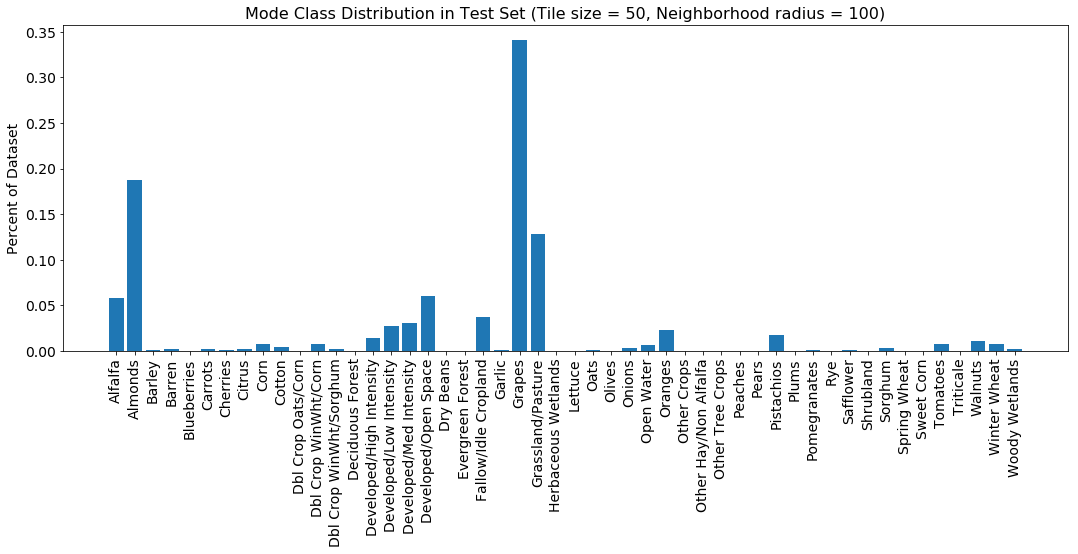}
  \caption{Distribution of test set tile labels for the NAIP dataset in Central Valley, CA. Here we show the distribution for tiles of size 50 and neighborhood radius 100.}
  \label{fig:testsetdist}
\end{figure}

The Cropland Data Layer (CDL) is a raster geo-referenced land cover map collected by the USDA for the entire continental United States \cite{cdl}. It is offered at 30 m resolution and includes 132 detailed class labels spanning field crops, tree crops, developed areas, forest, water, and more. In our dataset of Central Valley, we observe 66 CDL classes (Fig. \ref{fig:cdl} and Fig. \ref{fig:testsetdist}).

In this paper, we treat CDL labels as ground truth and use them to evaluate the quality of features learned by various unsupervised and supervised methods. To do this, we upsample CDL to NAIP resolution in Google Earth Engine, allowing us to assign every NAIP pixel a CDL class.
CDL is created yearly using imagery from Landsat 8 and the Disaster Monitoring Constellation (DMC) satellites, and a decision tree algorithm trained and validated on ground samples.
Our evaluation accuracies depend on the quality of the CDL labels themselves, which vary by class --- for the most common classes, accuracies detailed in the CDL metadata generally exceed 90\%.

\subsubsection{Landsat}
\label{appendix:landsat}

The Landsat satellites are a series of Earth-observing satellites jointly managed by the USGS and NASA. Landsat 8 provides moderate-resolution (30 m) satellite imagery in seven surface reflectance bands: ultra blue, blue, green, red, near infrared, shortwave infrared 1, and shortwave infrared 2 \citep{Roy:2014fla}. The spectral bands were designed to serve a wide range of scientific applications, from estimating vegetation biophysical properties to monitoring glacial runoff. Near-infrared and shortwave-infrared regions of the electromagnetic spectrum can capture ground properties that are difficult to see in the visible bands alone. For this reason, they are effective features in separating different land cover types, and often play a key role in classic, pixel-level supervised classification problems.

Landsat images are collected on a 16-day cycle and often affected by different type of contamination, such as clouds, snow, and shadows \citep{Whitcraft:2015co}. Although contaminated pixels can be removed using  masks that are delivered with the images, it can be challenging to obtain a completely clear image over large areas without human supervision. Remote sensing scientists often solve this problem by generating pixel-level composites of several images \citep{Luck:2016bj}. 

In this study, we generate and export median composites over three major US cities and their surrounding area: San Francisco, New York City, and Boston. Each image spans 1.2 degrees latitude and longitude and contains just under 20 million pixels.

We also gather Landsat 7 images from Uganda by bounding Uganda in a rectangular region and taking median composites from 2009-2011 using Google Earth Engine’s Landsat SimpleComposite tool. This region is then divided into $2^{14}$ patches of $\approx 145 \times 145$ pixels each of $20 \text{km}^2$. 

%\subsubsection{CIA Factbook} \label{appendix:data_factbook}

%\neal{Need to put in a table listing all features}

%%%%%%%%%%%%%%%%%%%%%%%%

\subsubsection{DigitalGlobe}
\label{appendix:dg}

DigitalGlobe is a company that provides high-resolution satellite data and supplies much of the imagery for Google Maps and Google Earth.
Image samples from a global composite with up to 0.3 m resolution are available free of charge through the Google Static Maps API.\footnote{https://developers.google.com/maps/documentation/static-maps/}

We sample a dataset of 11,564 satellite images tiling the area surrounding Addis Ababa, Ethiopia and spanning latitudes $[8.86, 9.15]$ and longitudes $[38.62, 38.91]$.
This dataset contains roughly 2.9 billion RGB pixels at zoom level 18, roughly corresponding to 0.6 m resolution.

%In contrast to the NAIP dataset, the DigitalGlobe imagery is captured from Earth-orbiting satellites rather than fixed-wing aircraft.

%%%%%%%%%%%%%%%%%%%%%%%%

\subsection{Experimental Details} 
\label{appendix:details}

All neural network-based approaches including Tile2Vec are implemented in PyTorch and trained with the Adam optimizer with learning rate 0.001 and betas $(0.5, 0.999)$.
ResNet-18 models are trained with batch size 50.
We use the scikit-learn random forest classifier implementation with 100 trees and default settings for all other parameters \cite{scikit-learn}.

Source code, trained models, and examples can be accessed at \url{https://github.com/ermongroup/tile2vec}.

% \newpage
% \section{OLD STUFF}

% \input{01_fig_supervised}

% \input{01_fig_loss}

\end{document}